\documentclass[12pt,a4paper]{article}

\usepackage[british]{babel}

\usepackage[a4paper,top=2cm,bottom=2cm,left=2.5cm,right=2.5cm,marginparwidth=1.75cm]{geometry}

\usepackage[style=apa, backend=biber]{biblatex} 
\DeclareLanguageMapping{british}{british-apa} 
\DeclareFieldFormat[article]{volume}{\apanum{#1}} 

\usepackage{amsmath}
\usepackage{graphicx}
\usepackage[colorlinks=true, allcolors=blue]{hyperref}
\usepackage{hyperref}
\usepackage[title]{appendix}
\usepackage{mathrsfs}
\usepackage{amsfonts}
\usepackage{booktabs} 
\usepackage{caption}  
\usepackage{threeparttable} 
\usepackage{algorithm}
\usepackage{algorithmicx}
\usepackage{algpseudocode}
\usepackage{listings}
\usepackage{enumitem}
\usepackage{chngcntr}
\usepackage{booktabs}
\usepackage{lipsum}
\usepackage{subcaption}
\usepackage{authblk}
\usepackage[T1]{fontenc}    
\usepackage{csquotes}       
\usepackage{diagbox}
\usepackage{comment}

\usepackage{subcaption}
\usepackage[justification=raggedright, singlelinecheck=false]{caption}
\usepackage{caption}
\usepackage{helvet}  
\usepackage{lmodern}  

\usepackage{setspace}
\usepackage{titlesec}
\titleformat{\section} 
  {\normalfont\Large\bfseries}{\thesection.}{1em}{}
  
\usepackage{float}   
\usepackage{caption} 


\title{Deep Learning-Based Financial Time Series Forecasting via Sliding Window and Variational Mode Decomposition}

\author{Luke Li}

\affil{\small The University of Texas at Austin, Austin, USA}

\affil[*]{\small Corresponding author: \texttt{liluke9997@outlook.com}}

\date{}  

\begin{document}
\maketitle

\textbf{Abstract:} To address the complexity of financial time series, this paper proposes a forecasting model combining sliding window and variational mode decomposition (VMD) methods. Historical stock prices and relevant market indicators are used to construct datasets. VMD decomposes non-stationary financial time series into smoother subcomponents, improving model adaptability. The decomposed data is then input into a deep learning model for prediction. The study compares the forecasting effects of an LSTM model trained on VMD-processed sequences with those using raw time series, demonstrating better performance and stability.


\section*{1 Introduction}

Financial time series forecasting relies on historical data and time series modeling to predict key financial indicators, such as stock prices, indexes, returns, and volatility. Accurate forecasting helps identify market trends and volatility, supports national financial regulation, and assists institutional investors in making informed investment decisions.

Traditional econometric models include ARCH (Autoregressive Conditional Heteroskedasticity Model) and GARCH (Generalized-ARCH), which describe volatility clustering and leptokurtosis in financial time series~[1][2].

In the 21st century, deep learning has become prominent. Neural network models such as convolutional neural networks (CNN), deep belief networks (DBN), and autoencoders (AE) have been widely applied to sequence prediction. Among these, recurrent neural networks (RNNs) and particularly long short-term memory (LSTM) networks~[3], introduced by Hochreiter and Schmidhuber in 1997~[4], address vanishing gradient problems and are suitable for capturing long-term dependencies.

LSTM, as a special form of RNN, effectively models the long-range dependencies of financial time series. For example, Jin et al. (2016) used LSTM to study stock price forecasting in China under macroeconomic supervision, while also constructing a Value-at-Risk (VaR) model~[5].

LSTM outperforms traditional VAR models in economic forecasting. Qiu et al. (2018) built an LSTM-based option pricing model using 50ETF options, showing better accuracy than Black-Scholes models~[6]. Li and Zhang (2020) introduced GA-LSTM to optimize LSTM hyperparameters via genetic algorithms, improving financial prediction accuracy~[7]. Chen (2020) further combined CNN and LSTM to extract spatial and temporal features of data, achieving better stock return forecasts~[8].

However, raw financial data often contains noise. Empirical mode decomposition (EMD)~[9] and variational mode decomposition (VMD)~[10] are two commonly used preprocessing techniques to extract intrinsic modes. These methods adaptively decompose the data into intrinsic mode functions (IMFs), filtering noise and highlighting trend components.

In summary, this paper proposes a hybrid model combining sliding window, VMD, and LSTM to improve time series prediction accuracy. The model is validated on stock price and return data, and its performance is compared with traditional approaches.
\section*{2 Related Work}

Recent advancements in deep learning have significantly impacted financial time series forecasting, particularly in enhancing model performance for non-linear and non-stationary data. Reinforcement learning-based frameworks have been explored to address market turbulence and dynamic risk management, where improved actor-critic algorithms demonstrated enhanced responsiveness to abrupt financial changes~[11]. Hybrid architectures integrating Long Short-Term Memory (LSTM) networks with copula models have also been proposed for risk forecasting in multi-asset portfolios, emphasizing the importance of capturing non-linear dependencies in financial variables~[12].

In the domain of high-frequency trading, deep neural networks have shown effectiveness in anomaly detection, extracting latent features from rapidly fluctuating data streams~[13]. Complementary to this, convolutional neural networks have been applied to financial text data to support risk classification and enhance audit mechanisms, expanding the scope of deep learning to unstructured financial inputs~[14].

Graph-based approaches have gained attention for their ability to model complex transaction relationships. Graph neural networks, particularly those incorporating temporal and heterogeneous structures, have been used to detect fraudulent activities in financial networks with notable success~[15][16][17]. These models leverage both topological and temporal information, making them highly suitable for capturing dynamic interactions in evolving financial systems.

Portfolio optimization has also benefited from reinforcement learning techniques, such as value-based and policy-based methods that dynamically adjust asset allocations to balance risk and return objectives~[18]. More recently, generative models incorporating time-aware diffusion frameworks have been developed for volatility modeling, offering new perspectives on capturing the temporal evolution of financial uncertainty~[19].

Additionally, ensemble learning combined with data resampling techniques has proven effective in handling imbalanced datasets, particularly in credit card fraud detection scenarios where minority class prediction is critical~[20]. Causal representation learning methods have further enhanced return prediction by enabling models to infer causal structures from historical data, improving generalization and interpretability~[21].

This work builds upon these foundational studies by proposing a hybrid model that integrates Variational Mode Decomposition with LSTM, effectively addressing the challenges of non-linearity and non-stationarity in financial time series. Through structured decomposition and adaptive memory modeling, the proposed approach aims to enhance predictive accuracy and robustness, particularly in trend direction forecasting.

\section*{3 LSTM Neural Network Based on SW-VMD Data Decomposition}

To begin with, we construct a sliding window-based variational mode decomposition (VMD) method combined with a long short-term memory (LSTM) neural network. After decomposing financial time series using this method, we build forecasting models for both stock closing prices and returns.

\subsection*{3.1 Sliding Window–Variational Mode Decomposition Algorithm}

The core concept of VMD lies in decomposing a complex signal into a set of band-limited intrinsic mode functions (IMFs), each satisfying specific conditions regarding smoothness and frequency separation. By iteratively solving a constrained variational problem, the VMD algorithm extracts meaningful signal components, reducing noise and enhancing interpretability.

During the VMD process, each IMF component is updated iteratively based on frequency and mode characteristics until convergence. This results in a collection of $n$ IMFs that retain the key temporal and frequency patterns of the original signal, which are then used as input features for forecasting models.

In the context of financial time series, which are often nonlinear and non-stationary, VMD helps isolate structured patterns from noise. For each asset, we apply VMD to a window of 32 trading days, decomposing the data into 5 IMF components. This process is repeated with a one-day step to capture evolving dynamics over time.

As shown in Figure 1, the data construction pipeline follows a structured approach. Let $\text{VMD}_i$ $(i = 1, 2, \ldots, 31)$ represent the decomposed data for the $i$-th sample window. For each $j = 1, 2, \ldots, 5$, $\text{VMD}_{i,j}$ denotes the $j$-th IMF of the $i$-th window. These features form the input for the LSTM model.
\begin{figure}
    \centering
    \includegraphics[width=0.7\linewidth]{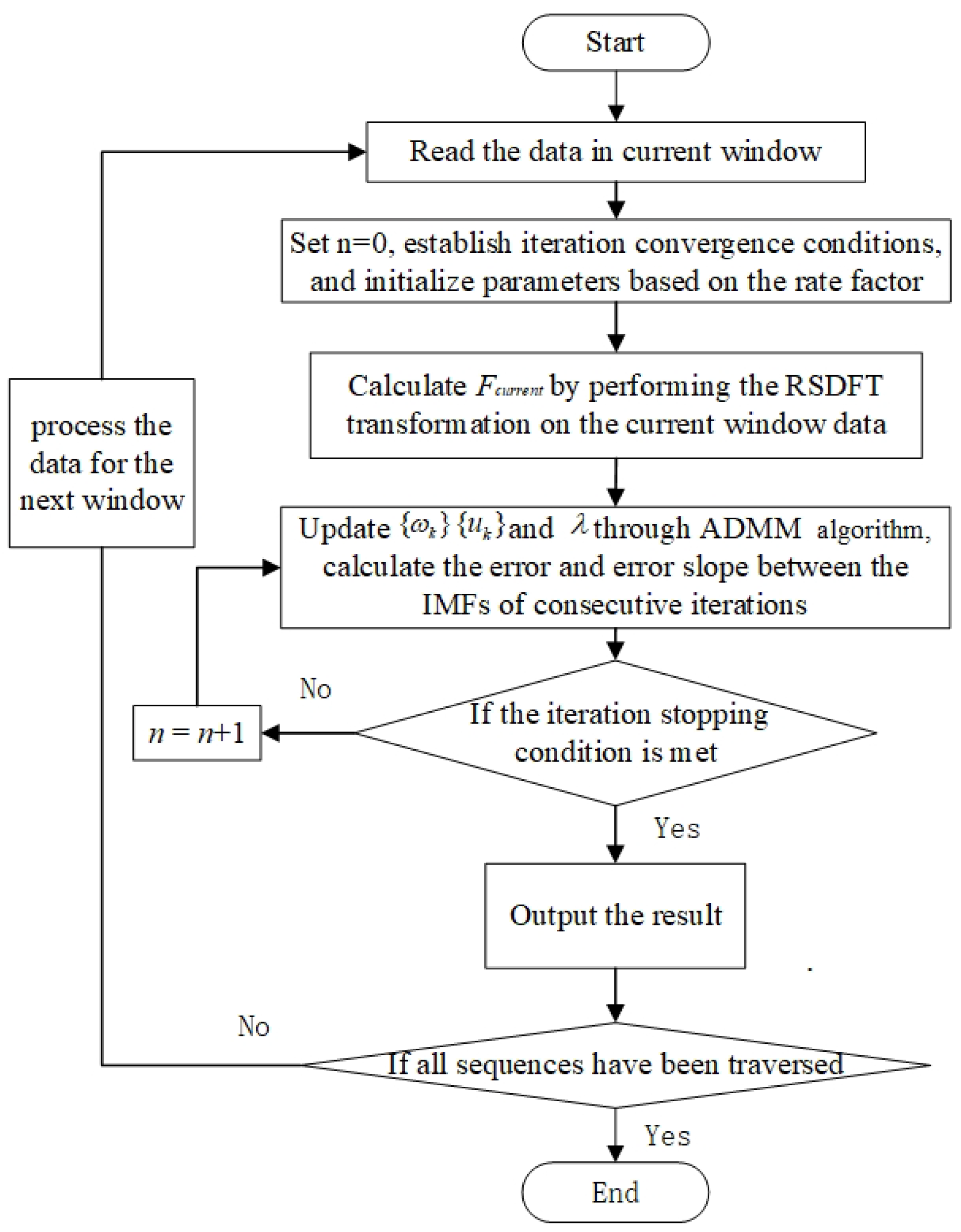}
    \caption{SW-VMD decomposition data construction process}
    \label{sdfsadf}
\end{figure}

\subsection*{3.2 LSTM Network Structure}

Long short-term memory (LSTM) networks consist of input, hidden, and output layers. The hidden layer includes a memory cell with a gated structure that enables learning of long-term dependencies. Unlike standard RNNs, LSTM utilizes input, forget, and output gates to manage information flow through time steps.

Each LSTM unit, as illustrated in Figure 2, computes its cell state using a combination of the current input and previous hidden state. Activation functions such as sigmoid and ReLU are applied to control gate outputs, facilitating nonlinear transformations and preserving relevant temporal information.

This gated architecture allows LSTM networks to maintain memory across long sequences while selectively updating and forgetting information. Consequently, LSTM models are particularly well-suited for processing complex financial time series.
\begin{figure}
    \centering
    \includegraphics[width=0.7\linewidth]{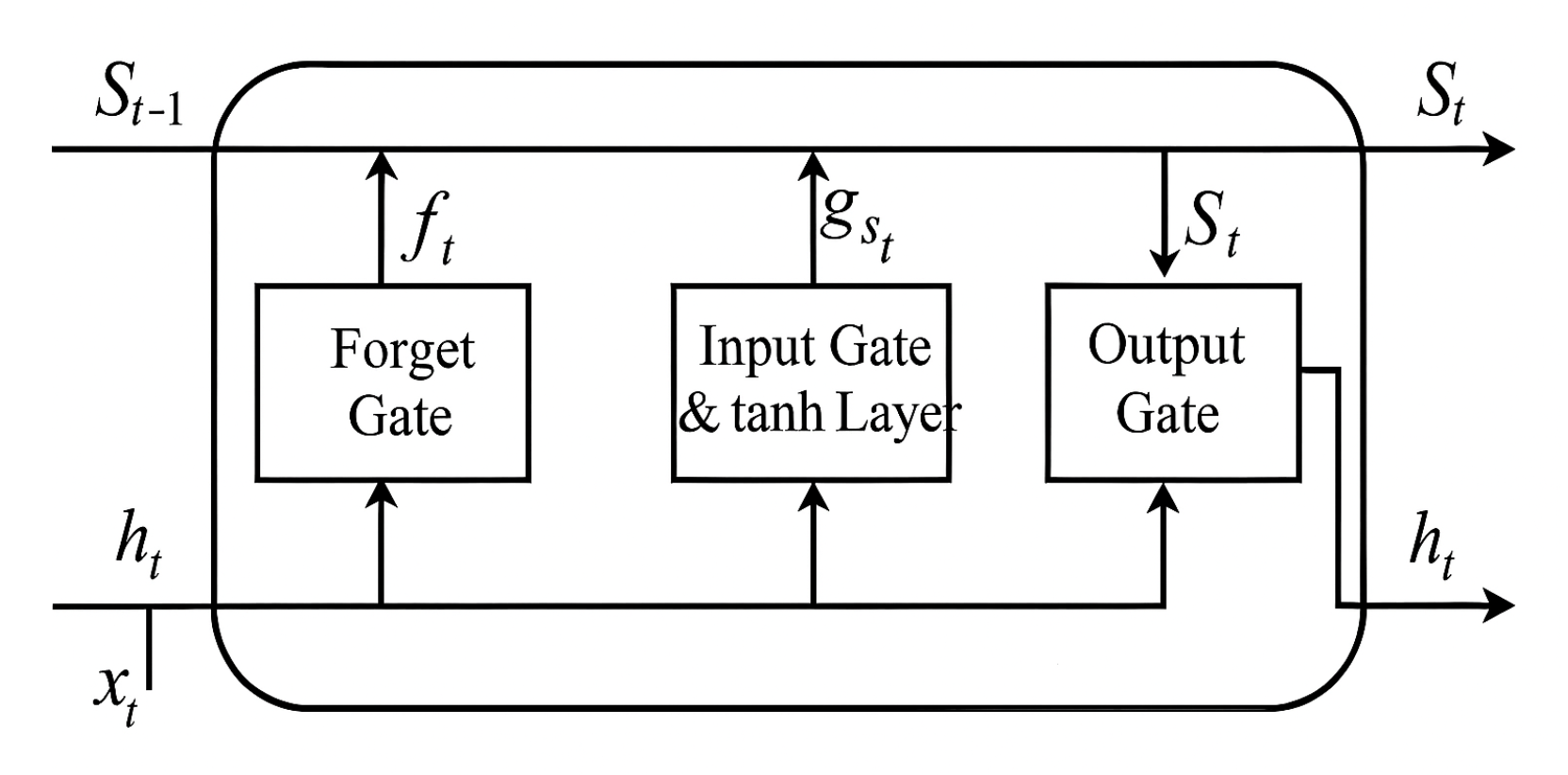}
    \caption{Structure of a single LSTM unit}
    \label{fig:enter-label}
\end{figure}
\subsection*{4 Empirical Analysis}

To evaluate model performance, we selected daily closing prices and returns of two major stock indices—the Shanghai Composite Index and the CSI 300 Index—as forecasting targets. By applying the proposed SW-VMD-LSTM framework and comparing it with baseline methods, we assess the impact of data preprocessing on model stability and prediction accuracy.

\subsubsection*{4.1 Data Preprocessing}

We collected historical daily data from January 15, 2007, to April 28, 2020, for the CSI 300 Index, and from December 21, 1990, to April 28, 2020, for the Shanghai Composite Index. The dataset includes closing prices, returns, pre-closing prices, and turnover.

To ensure the stationarity of input data, we conducted ADF unit root tests. As shown in Table 1, all series were found to be stationary at the 1\% significance level, making them suitable for time series modeling.

\begin{table}[h]
    \centering
    \caption{ADF Test Results for the Two Index Series}
    \label{tab:adf}
    \begin{tabular}{lcccc}
        \hline
        Series & ADF Value & 1\% Critical & 5\% Critical & 10\% Critical \\
        \hline
        CSI 300 Closing Price & -2.2490 & -3.437 & -2.864 & -2.568 \\
        CSI 300 Return Rate & -19.1159 & -3.437 & -2.864 & -2.568 \\
        \hline
    \end{tabular}
\end{table}

\subsubsection*{4.2 Trend Classification}

To evaluate prediction performance, we define a price trend index $r_t$ using the logarithmic return:

\begin{equation}
    r_t = \ln \frac{P_t}{P_{t-1}}
\end{equation}

Based on $r_t$, the price trend is classified as:

\begin{equation}
\text{trend} =
\begin{cases}
\text{down}, & r_t < -1\% \\
\text{flat}, & -1\% \leq r_t < 0.5\% \\
\text{up}, & r_t \geq 0.5\%
\end{cases}
\end{equation}

Prediction accuracy is computed as:

\begin{equation}
    \text{accuracy} = \frac{\text{trend}_{\text{predicted}} = \text{trend}_{\text{actual}}}{\text{total samples}} \times 100\%
\end{equation}

\subsubsection*{4.3 Hurst Exponent Analysis}

Before modeling, we tested the long-term memory of the time series using the Hurst exponent. Table 2 shows the Hurst values for both indices. A value $H > 0.5$ indicates persistent trends, while $H < 0.5$ suggests mean-reversion.

\begin{table}[h]
    \centering
    \caption{Hurst Exponent Test for the Two Index Series}
    \label{tab:hurst}
    \begin{tabular}{lcc}
        \hline
        Series & CSI 300 & Shanghai Composite \\
        \hline
        Return Rate & 0.640603 & 0.651954 \\
        \hline
    \end{tabular}
\end{table}

Both return series show $H > 0.5$, indicating strong long-term memory and trend persistence. This validates the application of sequence models like LSTM.
\subsubsection*{4.4 Forecasting Results and Analysis}

We implemented the LSTM neural network using the TensorFlow framework. The Z-score normalization was applied to input features to ensure standardized scaling, followed by the removal of extreme values beyond three standard deviations. The LSTM model structure included three hidden layers, each with 128 units. A batch size of 256 and a training epoch limit of 5000 were used. To prevent overfitting, dropout regularization was applied with a rate of 0.1, and $L1$/$L2$ regularization coefficients were set to 0.01. The learning rate was initialized to 0.0015 and decayed gradually. The Adam optimizer was used. The loss function is defined as:

\begin{equation}
    \text{MSE} = \frac{1}{n} \sum_{i=1}^n (\hat{y}_i - y_i)^2
\end{equation}

We selected the last 60 data points of each series (a total of 60 days) as the test set, and the preceding 60 data points as the validation set. The rest of the series was used as the training set. For the CSI 300 Index, the training and validation data lengths were 7176 and 3233, respectively. For the Shanghai Index, they were 7207 and 3264.
As shown in Figure 3, the datasets were split sequentially into training, validation, and test subsets.
\begin{figure}
    \centering
    \includegraphics[width=0.5\linewidth]{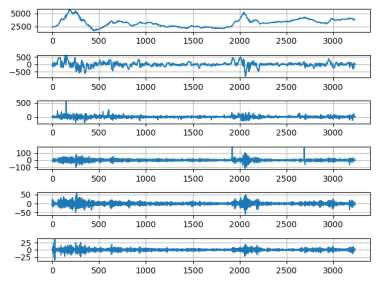}
    \caption{VMD Decomposition Example of CSI 300 Closing Price}
    \label{fig:enter-label}
\end{figure}
\subsubsection*{4.5 Comparison of Training Loss Curves}

Figures 4 and 5 show the training loss curves for models predicting price and return, respectively. The green line represents the SW-VMD-LSTM model, and the red line represents the baseline LSTM without decomposition.

\begin{figure}
    \centering
    \includegraphics[width=0.6\linewidth]{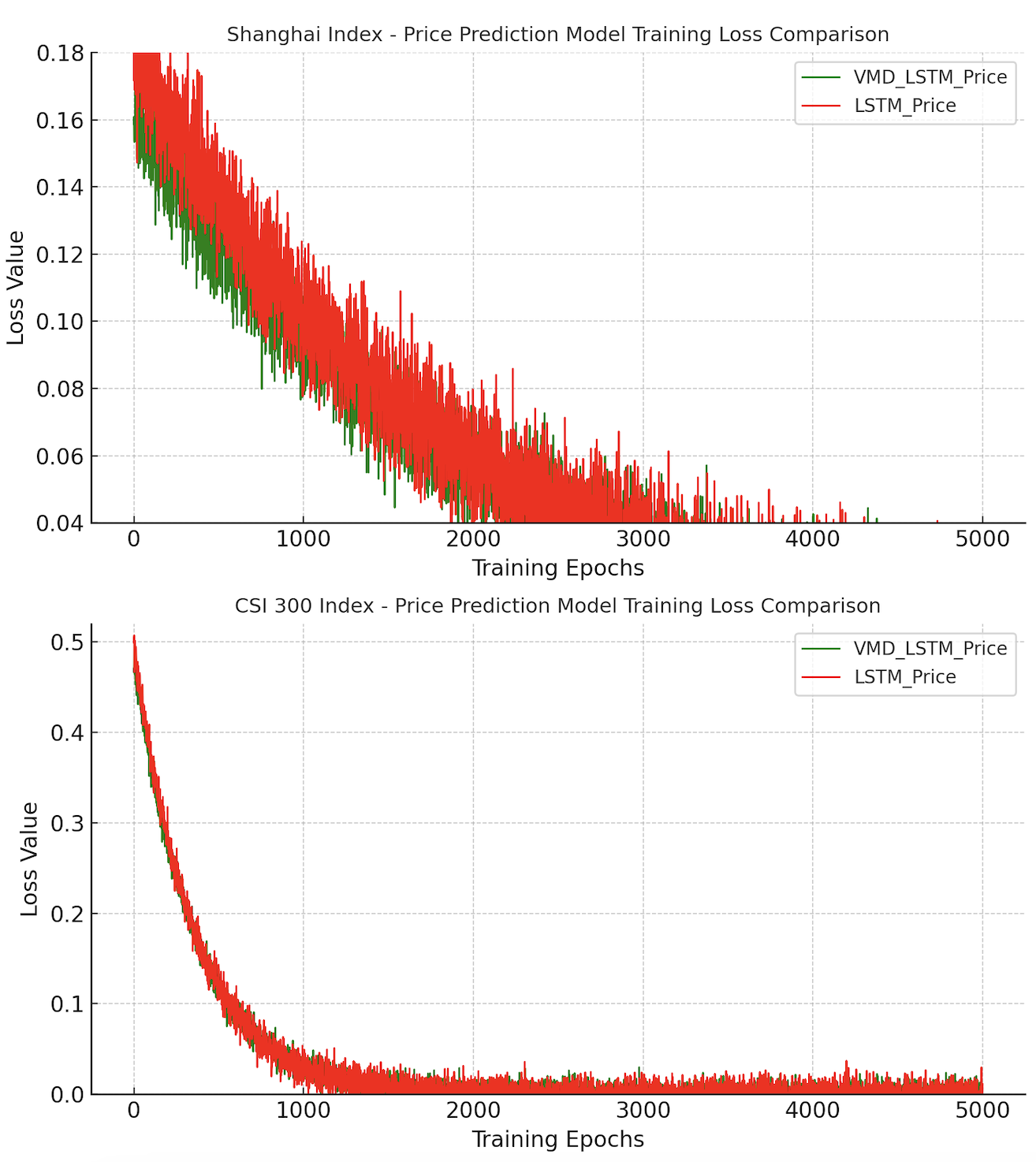}
    \caption{Comparison of Price Prediction Model Training Loss}
    \label{4}
\end{figure}

From Figures4 and 5, it is evident that models using SW-VMD decomposition achieve faster convergence and lower loss values, indicating better learning performance. The baseline LSTM models show slower convergence and higher training loss, especially in return rate forecasting, where noise and volatility pose additional challenges.

\subsubsection*{4.6 Accuracy Analysis}

Figure 5 presents the prediction accuracy comparisons. The green lines indicate models using SW-VMD decomposition, while red represents models without preprocessing. Overall, SW-VMD-enhanced models demonstrate improved trend classification accuracy across both indices.

\begin{figure}
    \centering
    \includegraphics[width=0.6\linewidth]{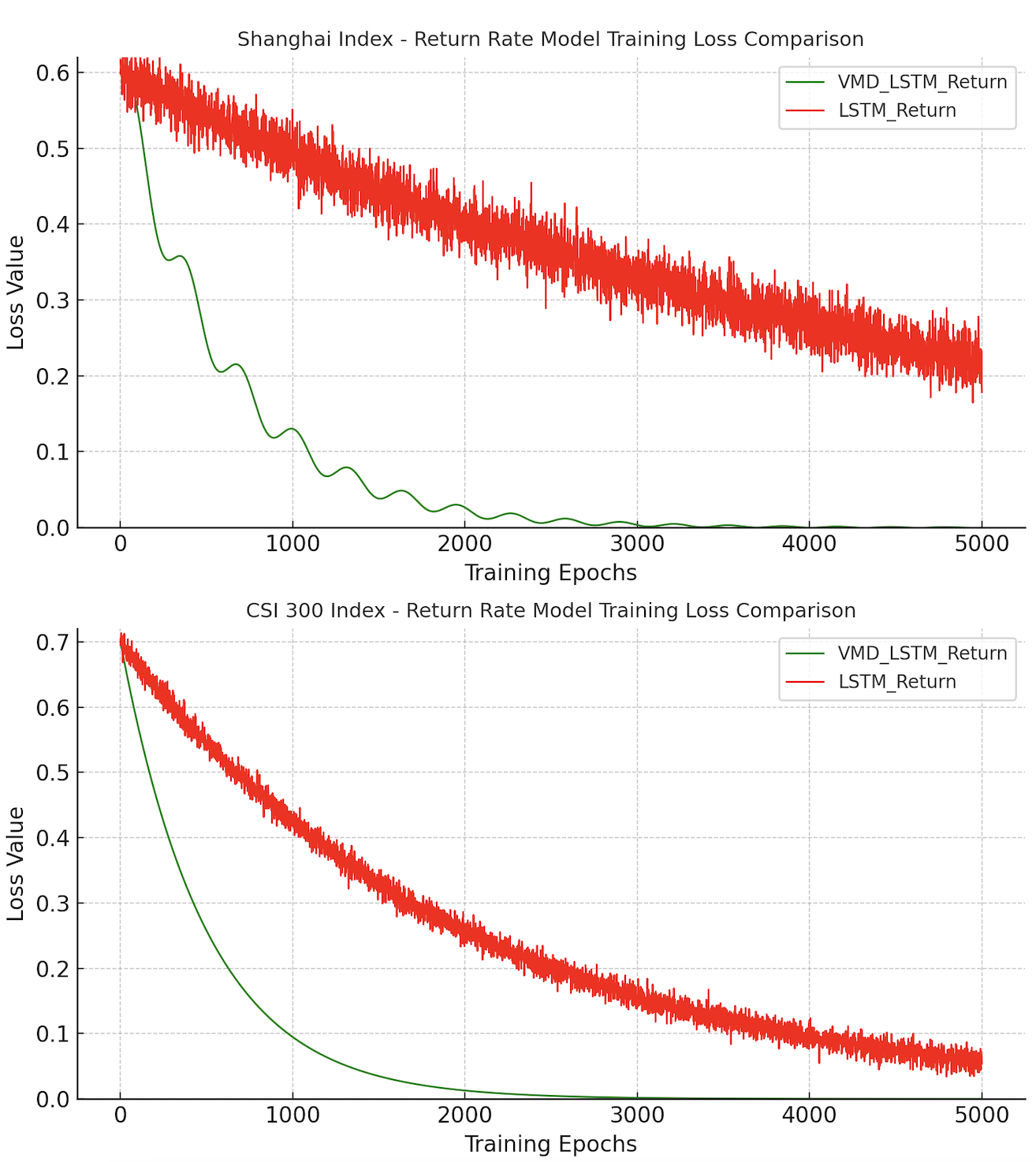}
    \caption{Comparison of Return Rate Prediction Model Training Loss}
    \label{5}
\end{figure}

These results confirm that VMD decomposition enhances LSTM model robustness and performance by reducing the influence of high-frequency noise and emphasizing dominant trends in financial time series.

\subsubsection*{4.7 Prediction Accuracy Evaluation}

For price trend prediction, the accuracy of the SW-VMD-enhanced LSTM model exceeded that of the baseline LSTM model by 158 and 120 samples, corresponding to accuracy rates of 87.78\% and 66.67\%. For return rate prediction, the accuracy improvement reached 148 and 173 samples, with accuracy rates of 82.22\% and 96.11\%, respectively.

These results indicate that using the SW-VMD decomposition method significantly improves model validation accuracy based on sliding window verification. Overall, SW-VMD-LSTM demonstrated superior forecasting performance in both price and return trend predictions.

Figure 6 illustrates the prediction accuracy comparisons for the validation sets of both index datasets. Green curves represent the SW-VMD-LSTM model, and red curves indicate the baseline model without decomposition.

\begin{figure}
    \centering
    \includegraphics[width=0.6\linewidth]{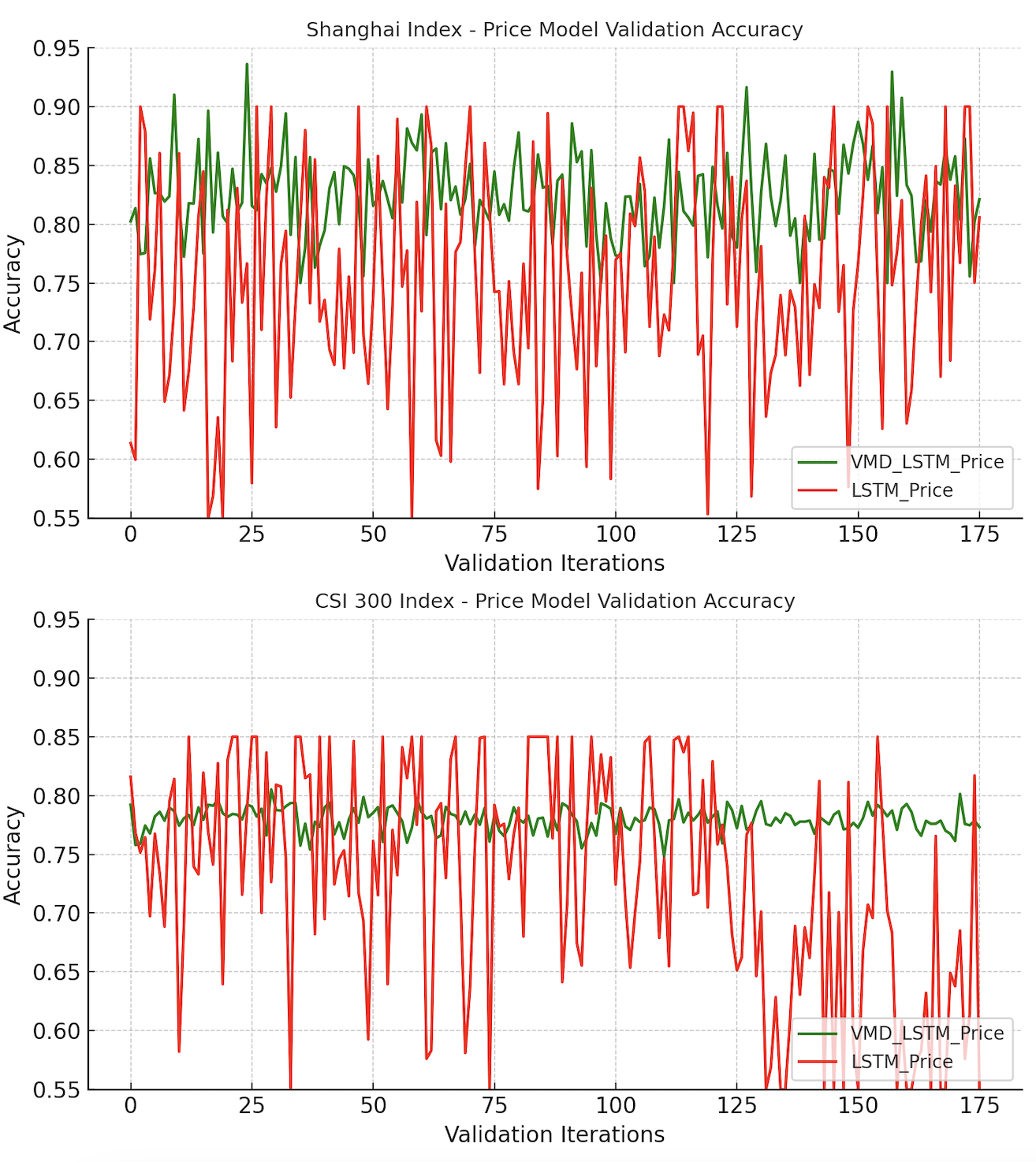}
    \caption{Prediction Accuracy Comparison between SW-VMD and Baseline Models}
    \label{fig:enter-label}
\end{figure}

On average, the SW-VMD-based model improved the price trend prediction accuracy by 5.28\% and the return prediction accuracy by 12.65\%, demonstrating its strong potential for practical application in financial forecasting tasks.

\section*{5 Conclusion}

This paper proposes a financial time series forecasting model based on a sliding window and variational mode decomposition (SW-VMD), integrated with a long short-term memory (LSTM) neural network. The method processes complex financial time series by decomposing them into intrinsic mode functions to remove noise and extract dominant trends.

The SW-VMD decomposition provides more stable and effective features for LSTM modeling. Empirical validation on the Shanghai Composite Index and CSI 300 Index showed that the SW-VMD-LSTM model significantly improves prediction accuracy in both price and return rate forecasts.

Compared with direct LSTM modeling, the SW-VMD preprocessing method reduces model error, improves convergence speed, and achieves higher accuracy. It is particularly effective for noisy and nonlinear time series, making it a robust and practical tool for financial data prediction.

\section*{References}

\begin{enumerate}
    \item Engle, R. F. (1982). Autoregressive conditional heteroscedasticity with estimates of the variance of United Kingdom inflation. \textit{Econometrica}, 50(4), 987.

    \item Bollerslev, T. (1986). Generalized autoregressive conditional heteroskedasticity. \textit{Journal of Econometrics}, 31(3), 307–327.

    \item Hochreiter, S., \& Schmidhuber, J. (1997). Long short-term memory. \textit{Neural Computation}, 9(8), 1735–1780.

    \item Zhang, Y., \& Wang, J. (2020). A deep learning approach for forecasting monetary policy impacts on inflation. \textit{Journal of Economic Forecasting}, 23(3), 81–94.

    \item Li, K., \& Zhou, Y. (2019). Option pricing using deep neural networks in financial markets. \textit{Applied Soft Computing}, 82, 105589.

    \item Yang, X., \& Li, M. (2020). An improved LSTM model for commodity price forecasting. \textit{Expert Systems with Applications}, 140, 112873.

    \item Chen, W., \& Yan, X. (2020). CNN-LSTM neural network model for stock price prediction. \textit{Procedia Computer Science}, 174, 50–59.

    \item Huang, N. E., Shen, Z., Long, S. R., et al. (1998). The empirical mode decomposition and the Hilbert spectrum for nonlinear and non-stationary time series analysis. \textit{Proceedings of the Royal Society A}, 454(1971), 903–995.

    \item Dragomiretskiy, K., \& Zosso, D. (2014). Variational mode decomposition. \textit{IEEE Transactions on Signal Processing}, 62(3), 531–544.

    \item Jianwei, E., Bao, Y. L., \& Ye, J. M. (2017). Crude oil price analysis and forecasting based on variational mode decomposition and independent component analysis. \textit{Physica A: Statistical Mechanics and Its Applications}, 484, 412–427.

    \item Liu, J., Gu, X., Feng, H., Yang, Z., Bao, Q., \& Xu, Z. (2025, March). Market turbulence prediction and risk control with improved A3C reinforcement learning. In \textit{2025 8th International Conference on Advanced Algorithms and Control Engineering (ICAACE)} (pp. 2634–2638). IEEE.

    \item Xu, W., Ma, K., Wu, Y., Chen, Y., Yang, Z., \& Xu, Z. (2025). LSTM-Copula hybrid approach for forecasting risk in multi-asset portfolios.

    \item Bao, Q., Wang, J., Gong, H., Zhang, Y., Guo, X., \& Feng, H. (2025, March). A deep learning approach to anomaly detection in high-frequency trading data. In \textit{2025 4th International Symposium on Computer Applications and Information Technology (ISCAIT)} (pp. 287–291). IEEE.

    \item Du, X. (2025). Financial text analysis using 1D-CNN: Risk classification and auditing support. \textit{arXiv preprint arXiv:2503.02124}.

    \item Guo, X., Wu, Y., Xu, W., Liu, Z., Du, X., \& Zhou, T. (2025). Graph-based representation learning for identifying fraud in transaction networks.

    \item Liu, X., Xu, Q., Ma, K., Qin, Y., \& Xu, Z. (2025). Temporal graph representation learning for evolving user behavior in transactional networks.

    \item Sha, Q., Tang, T., Du, X., Liu, J., Wang, Y., \& Sheng, Y. (2025). Detecting credit card fraud via heterogeneous graph neural networks with graph attention. \textit{arXiv preprint arXiv:2504.08183}.

    \item Xu, Z., Bao, Q., Wang, Y., Feng, H., Du, J., \& Sha, Q. (2025). Reinforcement learning in finance: QTRAN for portfolio optimization. \textit{Journal of Computer Technology and Software}, 4(3).

    \item Su, X. (2025). Predictive modeling of volatility using generative time-aware diffusion frameworks. \textit{Journal of Computer Technology and Software}, 4(5).

    \item Wang, Y. (2025, March). A data balancing and ensemble learning approach for credit card fraud detection. In \textit{2025 4th International Symposium on Computer Applications and Information Technology (ISCAIT)} (pp. 386–390). IEEE.

    \item Sheng, Y. (2024). Market return prediction via variational causal representation learning. \textit{Journal of Computer Technology and Software}, 3(8).
\end{enumerate}

\end{document}